\documentclass[letterpaper, 10 pt, conference]{ieeeconf}  

\IEEEoverridecommandlockouts                              

\usepackage{framed,multirow}
\usepackage{graphicx}
\usepackage{wrapfig}
\usepackage{multirow}
\usepackage{epstopdf}
\usepackage{tabulary}
\usepackage{hyperref}
\usepackage{mathptmx} 
\usepackage{times} 
\usepackage{amsmath} 
\usepackage{amsfonts}
\usepackage{amssymb}  
\usepackage{color}
\usepackage[bottom]{footmisc} 

\newcommand{\cblue}[1] {\textcolor{blue}{#1}}

\usepackage[T1]{fontenc} 

\begin{document}

\title{{\small \vspace{-1cm} \cblue{Accepted in IEEE/RSJ International Conference on
Intelligent Robots and Systems (\textbf{IROS 2019}), Macau, China }}\\ \vspace{0.5cm} Look Further to Recognize Better: Learning Shared Topics and Category-Specific Dictionaries for Open-Ended 3D Object Recognition}
\author{S. Hamidreza Kasaei
\thanks{Department of Artificial Intelligence, University of Groningen, PO Box 407, 9700 AK, Groningen, The Netherlands. Email: hamidreza.kasaei@rug.nl} 
\thanks{We are thankful to Prof. L. Seabra Lopes, Prof. A. Tom\'{e}, and Prof. M. Wiering who provided expertise that greatly assisted the research. 
}
}

\maketitle

\maketitle

\begin{abstract}
Service robots are expected to operate effectively in human-centric environments for long periods of time.
In such realistic scenarios, fine-grained object categorization is as important as basic-level object categorization. We tackle this problem by proposing an open-ended object recognition approach which concurrently learns both the object categories and the local features for encoding objects. In this work, each object is represented using a set of \emph{general latent visual topics} and \emph{category-specific dictionaries}. The general topics encode the common patterns of all categories, while the category-specific dictionary describes the content of each
category in details. The proposed approach discovers both sets of general and specific representations in an unsupervised fashion and updates them incrementally using new object views. Experimental results show that our approach yields significant improvements over the previous state-of-the-art approaches concerning scalability and object classification performance. Moreover, our approach demonstrates the capability of learning from very few training examples in a real-world setting. Regarding computation time, the best result was obtained with a Bag-of-Words method followed by a variant of the Latent Dirichlet Allocation approach.
\end{abstract}

\section{Introduction}

Nowadays service robots are leaving the structured and completely known environments and entering human-centric settings. For these robots, object perception is a challenging task due to the high demand for accurate and real-time response under changing and unpredictable environmental conditions.  Although many problems have already been understood and solved successfully, many challenges still remain. Open-ended object recognition is one of these challenges waiting for many improvements. Cognitive science revealed that humans learn to recognize object categories ceaselessly over time. This ability allows adapting to new environments, by enhancing their knowledge from the accumulation of experiences and the conceptualization of new object categories. Inspired by this, we approach object category learning and recognition from a long-term perspective and with emphasis on open-endedness. In this paper, \emph{open-ended} implies that the set of object categories to be learned is not known in advance, and the training instances are extracted from online experiences of a robot, and become gradually available over time, rather than being completely available at the beginning of the learning process. 

In such a complex, dynamic and realistic setting, no matter how extensive the training data used for batch learning, a robot might always face a new object. Therefore, apart from batch learning, the robot should be able to learn new object categories from very few training examples on-site supported by human-in-the-loop feedback. 
\begin{figure}[t]
\center
  \includegraphics[width=\linewidth, trim= 0.cm 0.cm 0.cm 0.3cm,clip=true]{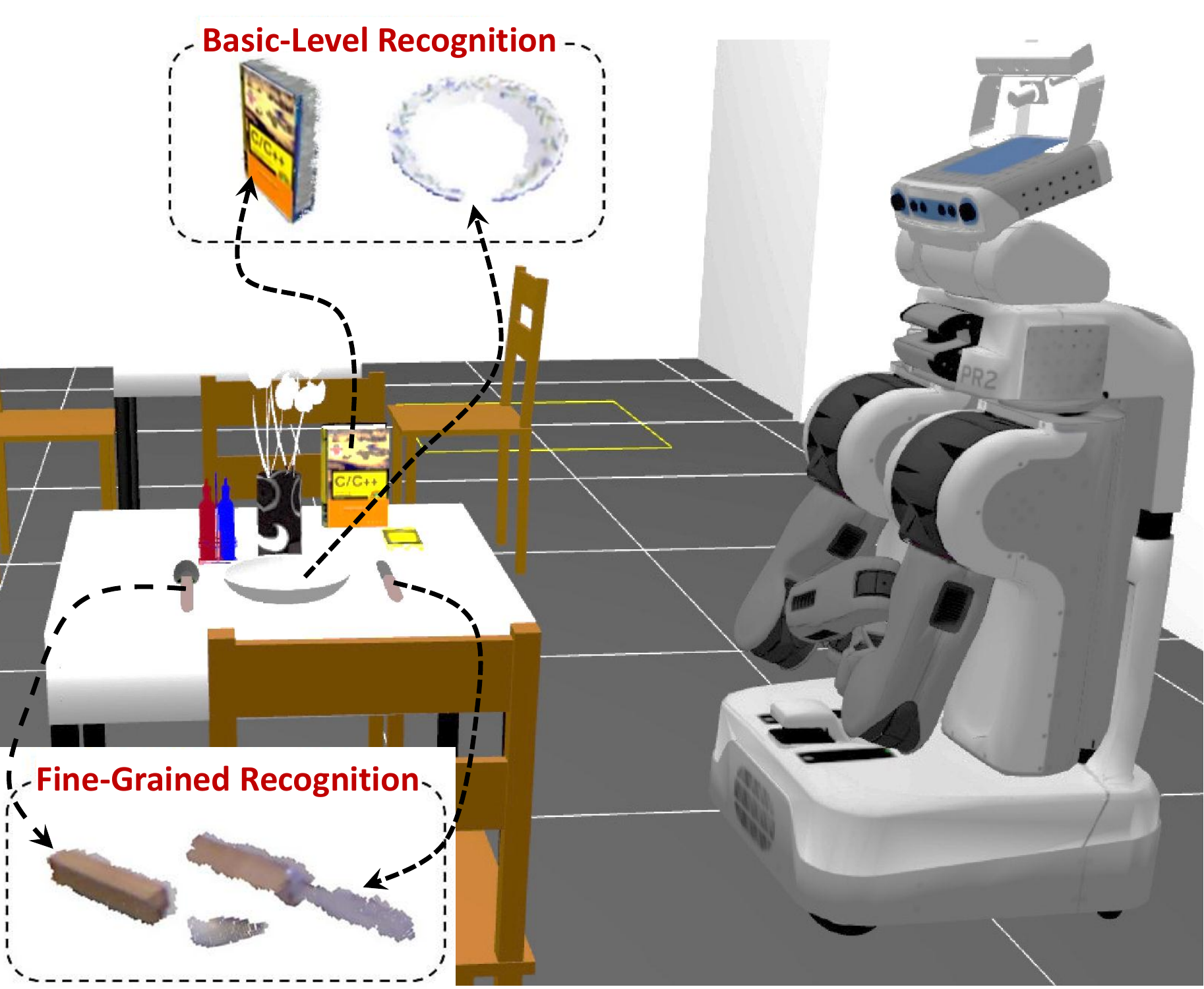}
\vspace{-6mm}
\caption{ An illustrative example of basic-level object categorization vs. fine-grained object categorization. PR2 robot is looking at some objects on the table. In such scenarios, distinguishing \emph{Spoon} from \emph{Knife} (fine-grained) is harder than distinguishing \emph{Book} from \emph{Plate} (basic-level) due to subtle differences in particular regions.}
\label{fig:fine-grained}       
\vspace{-5mm}
\end{figure}
Moreover, the robot may frequently face a new object that visually can be either \emph{not similar} (\emph{basic-level}) or \emph{very similar} (\emph{fine-grained}) to the previously learned object categories (see Fig.~\ref{fig:fine-grained}). This poses a significant challenge in situations when the robot needs to recognize visually similar categories for which only a few examples are available. In such situations, \emph{object representation} plays a central role because the output of this module is used for learning as well as recognition. In this paper, we propose a new method to characterize an object based on a \emph{common set of latent visual topics}, which is used to describe the content of all categories (basic-level), and a set of \emph{category-specific dictionaries} for highlighting the small diversity within the different categories (fine-grained). The representation of the given object is finally obtained by concatenating the generic and specific representations.  To the best of our knowledge, there is no other approach that can jointly learn a set of generic and category-specific features for encoding 3D object categories in an open-ended manner.

\section{Related work}
\label{sec:related_work}
In the last decade, various research groups have made substantial progress towards the development of learning approaches which support online and incremental object category learning~\cite{Oliveira2016614}\cite{young2017semantic}. In recent studies on object recognition, much attention has been given to deep Convolutional Neural Networks (CNNs). It is now clear that if in a scenario, we have \emph{a fixed set of object categories} and \emph{a massive number of examples per category that are sufficiently similar to the test images}, CNN-based approaches yield good results, notable recent works include \cite{redmon2016you}\cite{hariharan2017low}. In open-ended scenarios, these assumptions are not satisfied, and the robot needs to learn new concepts on-site using very few training examples. While deep learning is a very powerful and useful tool, there are several limitations to apply CNNs in open-ended domains. In general, CNN approaches are incremental by nature but not open-ended, since the inclusion of new categories enforces a restructuring in the topology of the network. Furthermore, training a CNN-based approach requires long training times and training with a few examples per category poses a challenge for these methods. 
In contrast, \emph{open-ended learning}~\cite{kasaeiNips2016}\cite{kasaei2015interactive} allows for concurrent learning and recognition. Our approach falls into this category. 

In the case of object representation, most of the recent approaches use either neural networks~\cite{ullrich2017selecting}\cite{hariharan2017low} or hand-crafted features~\cite{kasaei2018perceivingAAAI}. These approaches may not be the best option for such domains since the object representation procedure is a built-in component of the system. Oliveira et al.~\cite{oliveira2015concurrent} tackle this problem by proposing an approach for concurrent learning of visual codebooks and object categories in open-ended domains. Unlike our approach, they completely discard information related to the co-occurrence of local object features. Moreover, they did not consider fine-grained categorization. Existing approaches for fine-grained categorization heavily rely on accurate object parts/features annotations during the training phase. Such requirements prevent the wide usage of these methods. However, some works only use class labels and do not need exhaustive annotations. Geo et al.~\cite{gao2014learning} proposed a Bag of Words (BoW) approach for fine-grained image categorization. This work is similar to ours since they represent objects using generic and specific representations. However there are some differences: their codebooks are constructed offline; therefore, the representativeness of the training set becomes critical to the performance of the system.  Moreover, this approach is impractical for an open-ended domain (i.e., large number of categories) since the size of object representation is linearly dependent on the number of known categories. Zhang et al.~\cite{zhang2016weakly} proposed a novel fine-grained image categorization system. Similar to our work, they only used class labels during the training phase. Unlike our approach, in \cite{gao2014learning}\cite{zhang2016weakly} the representations of known categories do not change after the training stage. Moreover, they completely discarded co-occurrence (structural) information. This limitation might lead to non-discriminative object representations and, as a consequence, to poor object recognition performance. 

Several types of research have been performed to assess the added-value of structural information. Canini et al.~\cite{canini2009online} extended an online version of Latent Dirichlet Allocation (LDA) and proposed an incremental Gibbs sampler for LDA (here referred to as I-LDA). In online-LDA and I-LDA, the number of categories is fixed, while in our approach the number of categories is growing. 
Kasaei et al.~\cite{kasaeiNips2016} proposed an open-ended object category learning approach just by learning specific topics per category, while our approach does not only learn a set of general topics for basic-level categorization, but also learn a category-specific dictionary for fine-grained categorization.

\section {{Learning generic and category-specific representations }}
\label {sec:object_representation}

We organized our approach in three main phases: (\emph{i}) feature extraction; (\emph{ii}) learning generic representations; and (\emph{iii}) learning category-specific representations. In the following we describe each of these phases in detail.

\subsection{Feature Extraction}

\begin{figure}[!b]
\vspace{-5mm}
\center
\begin{tabular}[width=1\textwidth]{cccc}
 \includegraphics[width=0.2\linewidth, trim= 1cm 18.5cm 1.0cm 0cm,clip=true]{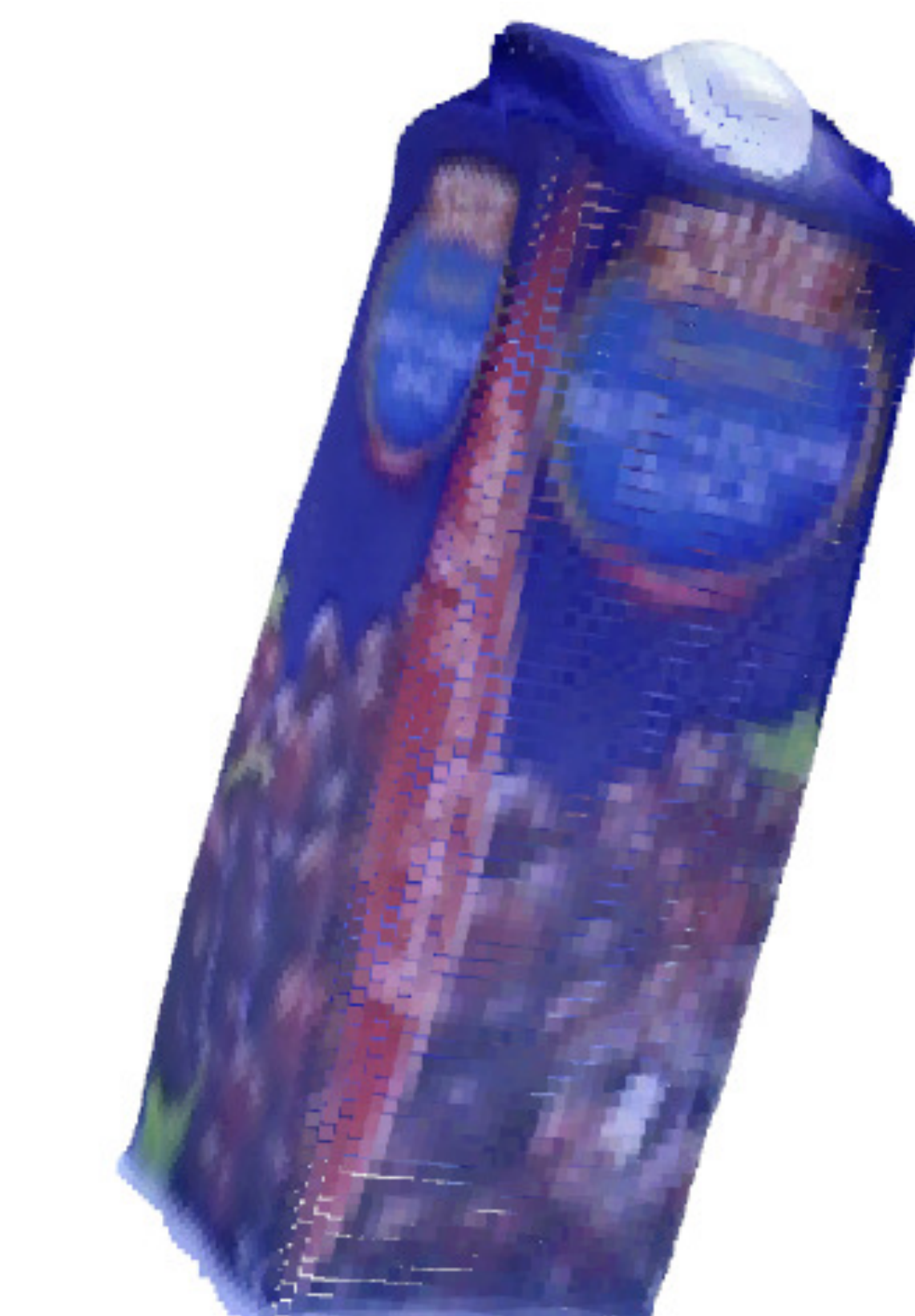} &
 \includegraphics[width=0.185\linewidth, trim= 1cm 1cm 1.0cm 17.5cm,clip=true]{figure/juice_box.pdf}&
\includegraphics[width=0.23\linewidth, trim= 0.5cm 0cm 0.0cm 0cm,clip=true]{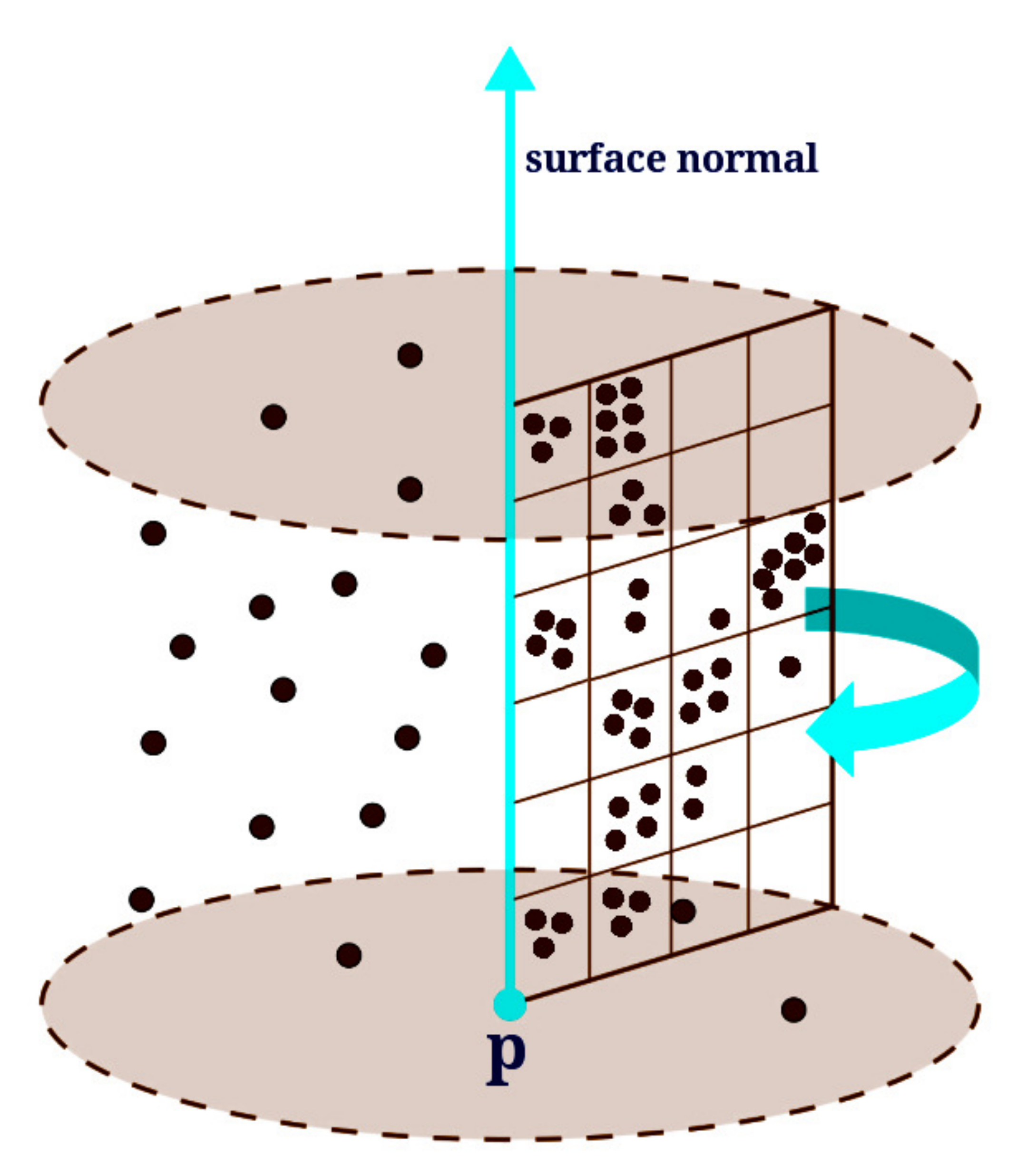}&
  \includegraphics[width=0.15\linewidth]{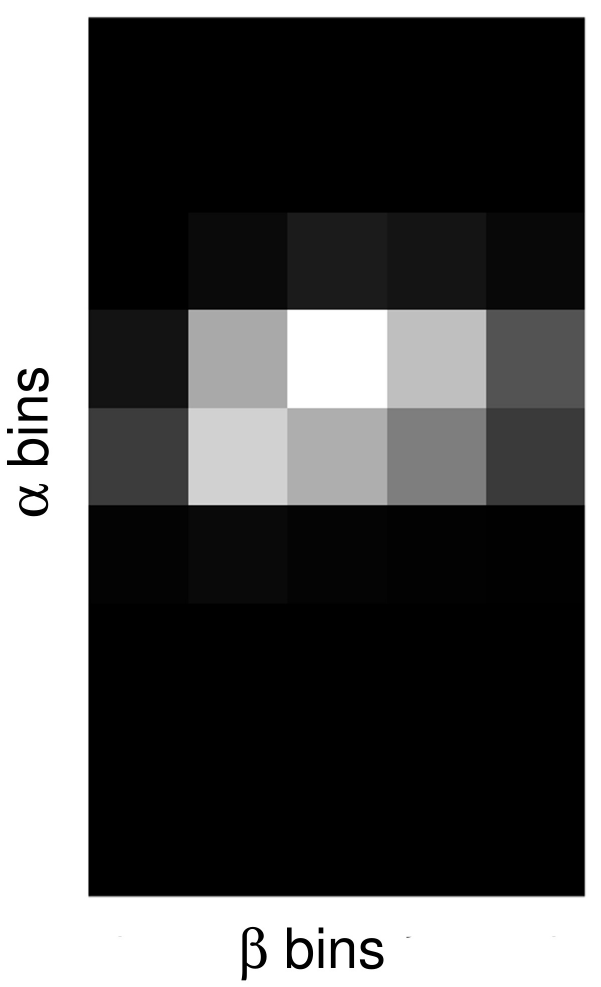} \\ 
 \emph{(a)} & \emph{(b)} &\emph{(c)} &\emph{(d)} \\
\end{tabular}\\
\caption{ Local feature extraction: \emph{(a)} point cloud of a juice-box \emph{(b)} key point selection (blue points); \emph{(c)} a schematic of how the spin-image spins a 2D histogram around the key point's surface normal; \emph{(d)} the computed spin-image for a sample keypoint.}
\label{fig:feature_extraction}       

\end{figure}

In this work, we first represent a 3D object by a set of local shape features called spin-image \cite{Johnson1999}. The reason why we use spin-image rather than other 3D feature descriptors is that the spin-image is a pose-invariant feature, and therefore suitable for 3D perception in autonomous robots. Another advantage of the spin-image is that it only requires a surface normal - rather than a full reference frame - to compute the feature. 
As depicted in Fig.~\ref{fig:feature_extraction}, the process of local feature extraction consists of two main phases: extraction of key points and computation of spin-images. For efficiency reasons, the number of key points in an object should be much smaller than the total number of points. Therefore, the object is first voxelized and then, the nearest neighbor point to each voxel center is selected as a key point (Fig.~\ref{fig:feature_extraction} \emph{a} and \emph{b}). Afterwards, the spin-image descriptor is used to encode the surrounding shape in each keypoint using the original point cloud. A spin-image is a local shape histogram, which is obtained by spinning a 2D histogram around the key point's surface normal (see Fig.~\ref{fig:feature_extraction} \emph{c} and \emph{d}). Therefore, each object view is described by a set of spin-images, $\textbf{O} = \{\textbf{s}_1,~\dots,~\textbf{s}_N\}$, where N is the number of key points. The obtained representation is then used as input to both generic and category-specific object representation processes.
\subsection {Generic Representation}

Our generic object representation approach requires a dictionary with $V$ visual words. Usually, the dictionary is created via off-line clustering of training data. In this work, we create a pool of objects using 75\% of the training data. Previously, we showed how a robot could create a pool of objects by exploring the environment~\cite{kasaeiNips2016}. 
To construct a pool of features, spin-images are computed for the selected points extracted from the pool of objects. Finally, the dictionary is constructed by clustering the features using the k-means algorithm~\cite{chakraborty2011analysis}. The centers of $V$ computed clusters are defined as the visual words, $\textbf{w}_i : i \in \{1,~\dots,~V \}$.

\begin{figure}[!t]
 \includegraphics[width=\linewidth, trim= 0cm 3cm 1cm 1.5cm,clip=true]{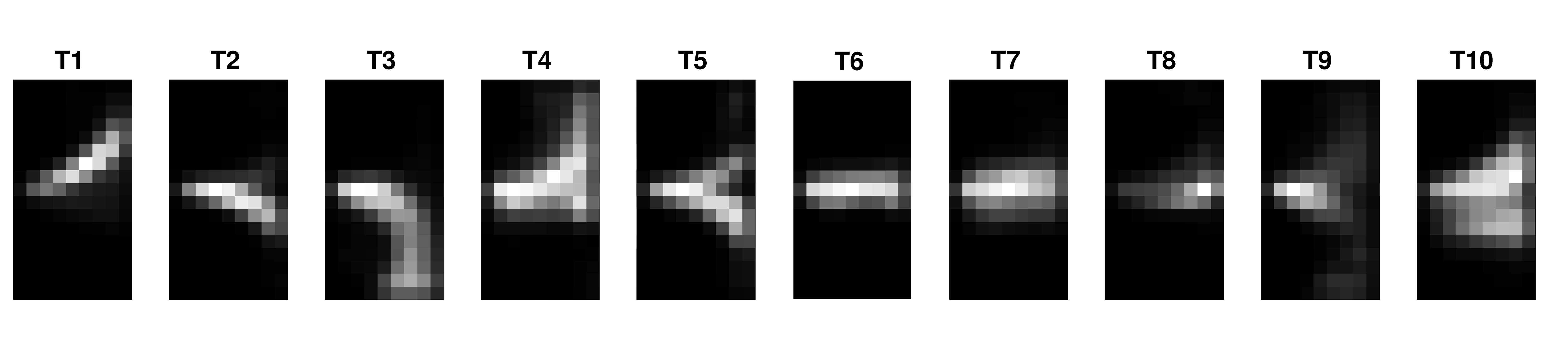}
 \vspace{-7mm}
\caption{\small A sample set of latent topics: spin-images compiled with IW~=~8~bins and SL~=~0.09m.}
\label{fig:topics}       
\vspace{-5mm}
\end{figure}

We then modify the incremental LDA approach~\cite{canini2009online} to be suitable for this study by releasing the batch learning phase. Particularly, we have tried to get structural semantic features from low-level feature co-occurrences by discovering a set of topics using an incremental LDA model. The basic idea is to represent a given object as a histogram of topics (i.e., latent variables), where a distribution over visual words characterizes each topic. 
In this method, an object is first represented as a set of visual words, $\{\textbf{s}_1,~\dots,~\textbf{s}_N\} \rightarrow \{{w}_1, {w}_2, ..., {w}_N\}$, where each entry represents the index of one of the $V$ words
of the dictionary. Next, the object should be described as a set of topics, $\{{w}_1, {w}_2, ..., {w}_N\}\rightarrow\{ {z}_1, {z}_2, ..., {z}_N\}$, where each element of the topic set represents the index of one of the $K$ topics. Towards this goal, the probability of topic $\textbf{z}_j$ being assigned to a word $\textbf{w}_i$, given all other topics assigned to all other words is estimated by the fast collapsed Gibbs sampler. After the sampling procedure, the word-topic matrix can be estimated as follows: 
\begin{figure}[!b]
\vspace{-4mm}
\center \includegraphics[width=0.95\linewidth, trim= 0.5cm 0cm 0.5cm 0.5cm,clip=true]{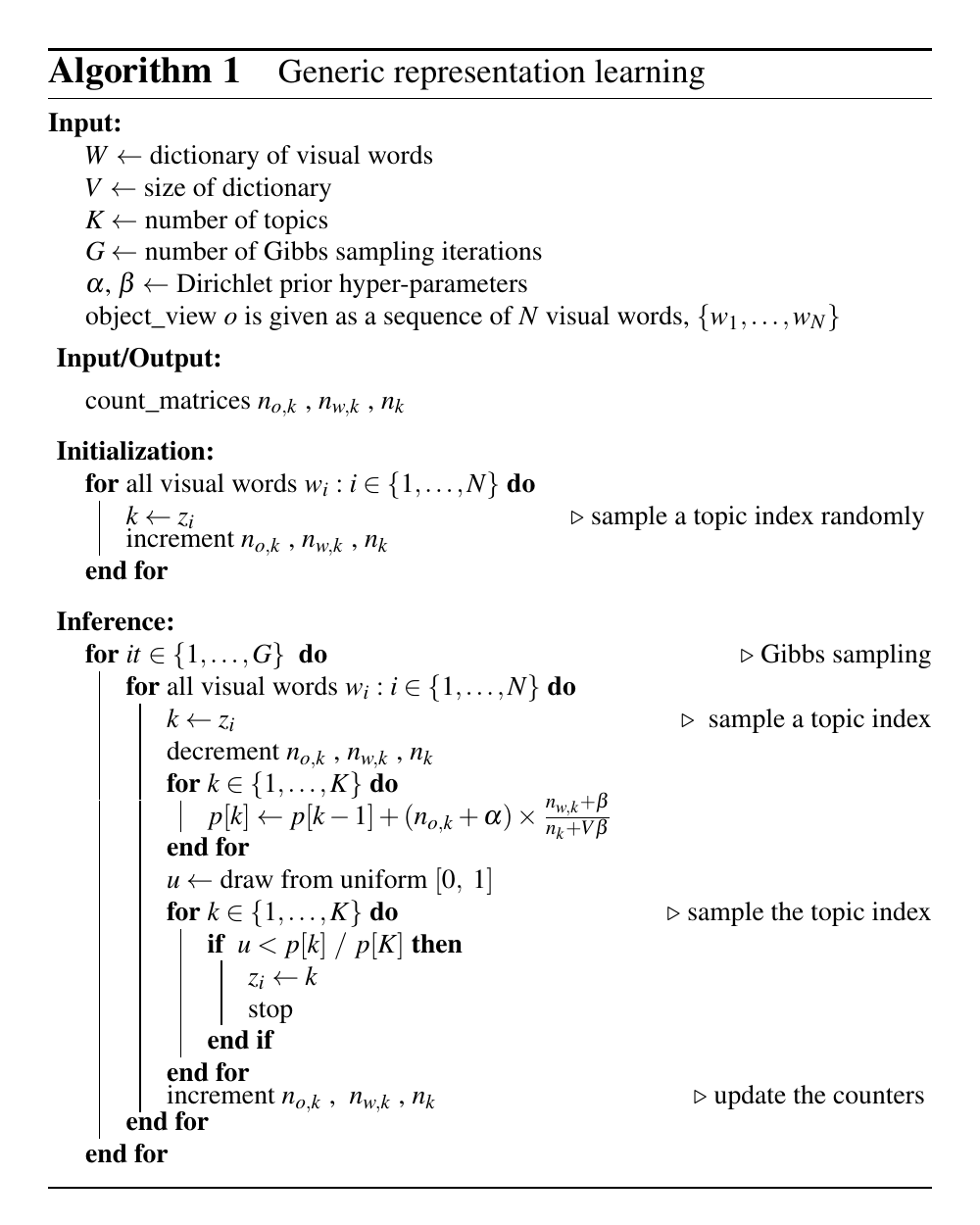}
 \vspace{-2mm}
\end{figure}
\begin{equation}
\label {phi_estimation}
\phi_{{w}_i,k} = \frac{n_{{w}_i, k} + \beta}{n_{k}+ |\textbf{V}|\times\beta},
\end{equation}
\noindent 
where $\beta$ is Dirichlet prior hyper-parameter that affect the sparsity of distributions, $n_{\textbf{w}_i,k} $ shows the number of times visual word $\textbf{w}_i$ was assigned to topic $k$ and $n_k$ is the number of times a word was assigned to topic $k$. $\phi$ is a $K \times V$ matrix, which represents words-probability for each topic, $K$ is the number of topics, and $\phi_{{w}_i,k}=p(\textbf{w}_i|\textbf{z}_k)$. As it is shown in Algorithm 1, we also need $n_{o,k}$ counter through the sampling procedure, which shows the number of times topic $k$ is assigned to some visual words of object $\textbf{O}$. After inferring the word-topic matrix, we generate a set of $K$ spin-image like topics:
\vspace{-2mm}
\begin{equation}
\label {topic_estimation}
\textbf{z}_k = \sum_{i=1}^{V} p(\textbf{w}_i|\textbf{z}_k) \times \textbf{w}_i \quad \quad k \in \{1, \dots, K\}.
\end{equation}
\noindent 
Each topic is then normalized to provide further robustness to depth variations of the object in the scene. In this way, each topic is generated by combining all visual words. It is worth mentioning that the process of topic learning does not require an explicit distance metric. The proposed procedure is summarized in Algorithm 1. At some points, a user may instruct the robot to update topics. In this case, the robot retrieves the representation of all the instances of all categories and updates dictionaries as well as the parameters of the model including $n_k$ and $n_{w, k}$ incrementally (i.e., unsupervised learning) using Gibbs sampling. Figure~\ref{fig:topics} shows a sample set of learned topics.
For representing a given object using the learned topics, each local feature of the object is approximated by its nearest topic. Then the object is represented as a histogram of occurrences of topics $\textbf{h}^t = [h_1,~ h_2,~\dots ,~h_K]$ where the $i^{th}$ element of $\textbf{h}^t$ is the count of the number of spin-images assigned to a topic, $\textbf{z}_i$.

\subsection {Category-Specific Representation}
While the generic representation works well for basic-level classification, it does not work for fine-grained categorization. The underlying reason is that most features from fine-grained categories are similar. Therefore, these categories would share lots of similar topics, and the proportion of discriminative topics would be minor. Therefore, it is desirable to develop an approach for learning a set of category-specific features from a few examples. We have tackled this problem by learning a category-specific dictionary for each category independently. This is a challenging task since the robot does not know in advance which object categories it will have to learn, which observations will be available, and when they will be available to support learning. In our current setup, a new instance of a specific category is stored in the robot's memory in the following situations:

\begin{itemize}
	  \item  When the teacher for the first time \emph{teaches} a particular category, through a \emph{teach} action, a new category is created and initialized with the set of object views.
       
    \item  In the case of \emph{correct} actions, the local features of the target object view are added to the category.

\end{itemize}
\noindent Assume at time $t_0$ (i.e., first teaching action) a dictionary is learned for category $c$, denoted as $V^c_{t_0}$, which represents the distribution of 3D shape features observed up to time $t_0$. Later at time $t_1$, a new training instance, which is represented as a set of spin-images, is taught by a teacher to category $c$ (i.e., supervised learning). The teaching instruction trigs the robot to retrieve the current dictionary of the category as well as the representation of the new object view and updates the relevant dictionary using an incremental K-means algorithm \cite{chakraborty2011analysis} (i.e. unsupervised learning). Such category-specific dictionary would highlight the differences of objects from different categories, and as a consequence improves the object recognition performance.

\begin{figure}[!t]
 \includegraphics[width=1\linewidth, trim= 0cm 3cm 0cm 1.5cm,clip=true]{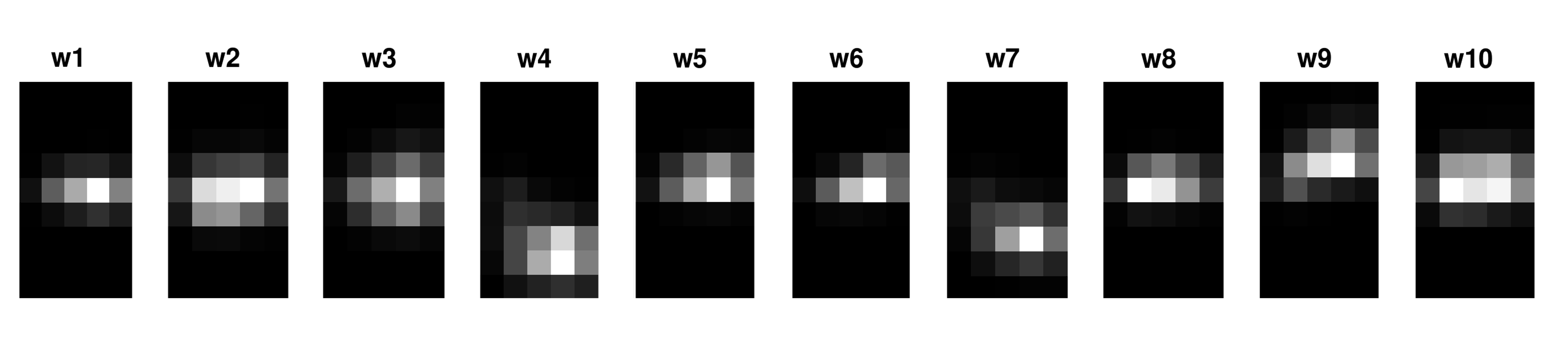}
  \vspace{-7mm}
\caption{\small A sample set of visual words of a category-specific dictionary: spin-images compiled with IW~=~4~bins and SL~=~0.09m.}
\label{fig:visual_words}       
\vspace{-5mm}
\end{figure}

Similar to the generic representation, a given object is then represented as a histogram of occurrences of visual words, $\textbf{h}^{c} = [h_1,~ h_2,~\dots ,~h_{V^c}]$, where $V^c$ is the size of category-specific dictionary. The obtained histograms, $\textbf{h}^t$ and $\textbf{h}^{c}$, are then concatenated to form a single representation for the given object. Figure \ref{fig:visual_words} shows ten sample visual words from a \emph{Mug} category-specific dictionary. It is worth mentioning when dictionaries or topics are updated, the representations of known instances must be updated. At the moment, we do this by storing the features of each object view and using them to recompute the representations. 

\begin{table*}[!t]
\begin{center}
\vspace{-1mm}
\caption {\small Object recognition performance for different parameters using restaurant object dataset.}
\resizebox{\linewidth}{!}{
\begin{tabular}{ |c|c|c|c|c|c|c|c|c|c|c|c|c|c|c|c|c|c|c|c|c|c| }
\hline
Parameters & \multicolumn{2}{|c|}{VS (m)} &\multicolumn{2}{c}{IW (bins)} &
\multicolumn{3}{|c|}{SL (m)} &	\multicolumn{5}{|c|}{V (generic words)} &\multicolumn{3}{|c|}{K (topics)} &	\multicolumn{5}{|c|}{$\operatorname{V}^c$ (specific words)} \\
\hline
Values & 0.02 & 0.04 & 4 & 8 & 0.07 & 0.08 & 0.09 & 50 & 60 & 70 & 80 & 90 & 70 & 80 & 90 & 50 & 60 & 70 & 80 & 90 \\
\hline
Avg. Accuracy(\%) & \textbf{85} & 81 & 83 & \textbf{83} & 82 & 83 & \textbf{83} & 82 & 82& 83& 84& \textbf{84}& \textbf{84} & 83 & 82 & 81& 84& \textbf{84} & 83 & 82\\
\hline
\end{tabular}
}
\vspace{-6mm}
\label{table:system_params_topic_words}
\end{center}
\end{table*}

\section {Object category learning and recognition}

Concerning category formation, we use the instance-based learning and recognition (IBL) approach which considers category learning as a process of learning about the instances of the category, i.e., a category is represented simply by a set of known instances. IBL is a baseline approach to evaluate object representations. An advantage of the IBL approaches is that they can recognize objects using a very small number of experiments and the training phase is very fast. As discussed in the previous section, the \emph{teach} and \emph{correct} actions by the teacher lead the robot to create a new category or to modify an existing one (i.e., supervised learning). Whenever a new object is added to a category, the agent retrieves the current model of the category and updates the category model by storing the representation of new object views. In particular, our approach can be seen as a combination of a particular \emph{object representation}, \emph{similarity measure} and \emph{classification rule}.  Therefore, the choice of the similarity metric has an impact on the recognition performance. 
 
With regards to the similarity measure, since the proposed object representation describes an object as a histogram, the dissimilarity between two histograms can be computed by different distance functions. We refer the reader to a comprehensive survey on distance/similarity measures provided by Cha~\cite{cha2007comprehensive}. After performing several cross-validation experiments, we concluded that two types of distance functions including Jensen-Shannon (JS) and chi-squared ($\chi^{2}$) distances are suitable to estimate the similarity between two instances. Both functions are in the form of a bin-to-bin distance function. Although the practical results of $\chi^{2}$ and JS are almost identical, $\chi^{2}$ is computationally more efficient. Therefore, we use the $\chi^{2}$ function to estimate the similarity of two instances.
Mathematically, let $\operatorname{P}$ and $\operatorname{Q}$ $\in{\rm I\!R^{K+V^c}}$ be the representation of two objects:
\begin{equation}
\chi^{2}\operatorname{(P,Q)} = \frac{1}{2}\sum_{i=1}^{K+V^c} \frac{(\operatorname{P}_i-\operatorname{Q}_i)^2}{(\operatorname{P}_i+\operatorname{Q}_i)}.
\end{equation}

To assess the dissimilarity between a target object and stored instances of a certain category \textbf{C}, the target object should be first described by the general topics and the learned dictionary of the category \textbf{C}. Afterwards, the minimum distance between the target object and all stored instances of the category \textbf{C} is considered as the Object-Category-Distance (OCD). The target object is finally classified based on the minimum OCD.
 
\section{Result and discussion}
\label{sec:results}
Three types of experiments were carried out to evaluate the proposed approach. In this section, we first explain our experimental setup and then discuss the obtained results. 
\subsection {Datasets and Baselines}
The experimental evaluations were carried out on two 3D object datasets including the Restaurant Object dataset~\cite{kasaei2015interactive} and the Washington RGB-D Object dataset~\cite{Lai2011}. 
For the classical evaluation (i.e., 10-fold cross-validation), we mainly use the Restaurant Object Dataset since it has a small number of classes (10 categories) with significant intra-class variation that is suitable for performing extensive sets of experiments. The Washington RGB-D Object dataset~\cite{Lai2011} is used for open-ended evaluation. It consists of 250,000 views of 300 common household objects taken from multiple views and organized into 51 categories. We also report on a real-world demonstration using the Imperial College Domestic Environment Dataset~\cite{doumanoglou2016recovering}. This  is a suitable dataset for this test since all scenes were captured under various clutter and contain objects with similar shapes (\emph{lipton} vs. \emph{softkings}) and objects with very different shapes (\emph{oreo} vs. \emph{elite}). 
 
For comparison, we have selected four open-ended 3D object category learning and recognition approaches, including the RACE~\cite{Oliveira2016614}, BoW~\cite{kasaei2018towards} based on the nearest neighbour classification rule, and Open-Ended LDA, which is a modified version of the standard smoothed LDA~\cite{blei2003latent} and Local-LDA~\cite{kasaeiNips2016}. Moreover, we add another baseline, which is the proposed method without category-specific representation (here referred to as Generic Rep.).
\vspace{-1mm}
\subsection {Classical Evaluation using Restaurant Object Dataset}
The proposed approach has a set of parameters including $\langle\alpha$, $\beta$, G, VS, IW, SL, V, K and V$^c\rangle$, that must be tuned to provide a good balance between recognition performance, memory usage, and processing speed. In this work, we assumed a symmetric Dirichlet prior for both $\alpha$ and $\beta$ parameters. Therefore, a high $\alpha$ value means that each object is likely to contain a mixture of most of the topics, and not a single specific topic. Likewise, a low $\beta$ value means that a topic may contain a mixture of just a few of the words. We carried out several experiments and concluded that $\alpha$ and $\beta$ should be set to 1.0 and 0.1 respectively. The number of Gibbs sampling iterations, G, is set to 50. A set of 900 experiments was performed for different values of the remaining six parameters. The voxel size (VS) parameter is related to the number of keypoints extracted from each object view. IW defines the size of the spin-image descriptor, which will be $(\operatorname{IW} + 1) \times (2 \times \operatorname{IW} + 1)$ float (4 bytes). Support length (SL) determines the amount of space swept out by a spin-image. A summary of the experiments using Restaurant Object dataset~\cite{kasaei2015interactive} is reported in Table~\ref{table:system_params_topic_words}. The parameter configuration that obtained the best \emph{average accuracy} was selected as the default configuration (i.e., bold numbers). The accuracy of this configuration was 94$\%$.  
A complete experiment (including both learning and recognition phases) on average took 286.50 seconds. 

\subsubsection{Comparative evaluation} 
Table~\ref{table:compareWithStateOfTheArt} presents a summary of the obtained results. One important observation is that the overall performance of our approach is promising and the proposed representation is capable of providing a distinctive representation for the objects. Moreover, it was observed that the discriminative power of our approach was better than the other evaluated approaches which was 7 percentage points (p.p.) better than RACE, 5 p.p. better than BoW, 11 p.p. and 3 p.p. better than Open-Ended LDA (shared topics) and Local LDA (category-specific topics) respectively. Furthermore, to show the importance of considering the category-specific representation, we compared our approach with the Only Generic Rep. baseline. It was observed that the accuracy has been boosted by 4 p.p. when we use both generic and category-specific representations. 
The accuracy of object recognition based on variable size representation (i.e., RACE) was not as good as the other approaches. The Local LDA and BoW obtained an acceptable performance. In the case of experiment times, BoW achieves the best performance and RACE was the most computationally expensive approach.  Overall, topic modelling approaches achieve a medium-level 
\begin{wraptable}{r}{0.5\linewidth}
\begin{center}
\vspace{-3mm}
\caption {\small Recognition performance}
\vspace{-5mm}
\resizebox{\linewidth}{!}{
\begin{tabular}[t]{|c|c|c|}
  \hline
  \textbf{Approach} & \textbf{Accuracy} & \textbf{Time}~(s)\\
  \hline
   RACE\cite{Oliveira2016614} & 0.87 & 1757\\ \hline
   BoW\cite{kasaei2018towards} & 0.89 & \textbf{196}\\\hline
   LDA\cite{blei2003latent}  &  0.83 & 258\\\hline
   Local-LDA\cite{kasaeiNips2016} &  0.91 & 349\\ \hline
   Generic Rep. &  0.90 & 234\\ \hline
   \textbf{Our Work} &  \textbf{0.94} & 287 \\ 	
 \hline
\end{tabular}}
\vspace{-7mm}
\label{table:compareWithStateOfTheArt}
\end{center}
\end{wraptable}
experiment time. 
The underlying reason is that there is a Gibbs sampling procedure in the topic modelling based approaches which takes time to accurately represent the desired distribution.

\subsection {Open-Ended Evaluation}
An evaluation protocol for open-ended learning systems was proposed in~\cite{kasaei2018coping}. The idea is to emulate the interactions of a robot with the surrounding environment over large periods of time. This protocol is based on a Test-then-Train scheme.
We developed a \emph{simulated teacher} to follow the protocol and autonomously interact with the agen.  The idea is that for each newly taught category, the simulated teacher repeatedly picks unseen objects of the currently known categories from a dataset and presents them to the agent. It progressively estimates the recognition accuracy of the agent and, in case this accuracy exceeds a given threshold ($\tau$ = 0.67, meaning accuracy is at least twice the error rate), introduces an additional object category. This way, the agent is trained online, and at the same time, the accuracy of the system is continuously estimated. In case the agent can not reach the classification threshold after a certain number of iterations (i.e., 100 iterations), the simulated teacher can infer that the agent is no longer able to learn more categories and terminates the experiment. We assess our experimental results using the metrics that were recently introduced in~\cite{oliveira2015concurrent}, including: (\emph{i}) the average number of
learned categories at the end of an experiment~(ALC), an indicator of \emph{how much the system is capable of learning}; (\emph{ii}) the number of question/correction iterations (QCI) required to learn those categories and the average number of stored instances per category (AIC), indicators of \emph{how much memory does it take for learning?}; (\emph{iii}) Global Classification Accuracy (GCA), an accuracy computed using all predictions in an experiment, and the Average Protocol Accuracy (APA), indicators of \emph{how well the system learns}.

\subsubsection{Results} 
A detailed summary of the obtained results is reported in Table~\ref{table_open_ended_evaluation}. In all approaches, as more categories are learned, the classification accuracy first decreases (performance degradation phase), and then starts going up again as more instances are introduced (recovery phase). Eventually, the agent reaches a break-point where it is no longer able to learn more categories.
By comparing all approaches, it is visible that the agent learned (on average) more categories using our approach than with other state-of-the-art approaches. The agent with Local-LDA approach obtained acceptable scalability, while the scalability of the other approaches was very low and their performance drops aggressively when the number of categories increases. In particular, our approach on average learned around $11$ categories more than Local-LDA and $18$, $20$ and $25$ categories more than BoW, RACE and open-ended LDA approaches, respectively. It is also clear that the agent with our approach stored more instances per category (AIC) than the other approaches. This is expected since a higher number of categories known by the system tends to make the classification task more difficult. 
\begin{table}[!t]
\begin{center}
\caption {Summary of open-ended evaluation.}
\resizebox{\columnwidth}{!}{
\begin{tabular}{ |c|c|c|c|c|c| }
\hline
\textbf{Approaches} & \#$\operatorname{\textbf{QCI}}$ & $\operatorname{\textbf{ALC}}$ & $\operatorname{\textbf{AIC}}$ & $\operatorname{\textbf{GCA}}$ & $\operatorname{\textbf{APA}}$ \\
\hline \hline
RACE\cite{Oliveira2016614} &382.10 &19.90 & 8.88 & 0.67 & 0.78\\
\hline 
BoW\cite{kasaei2018towards}& 411.80 &21.80 & 8.20 & 0.71 & 0.82\\
\hline 
 LDA\cite{blei2003latent} & 262.60 &14.40 & 9.14& 0.66& 0.80\\
\hline
Local-LDA\cite{kasaeiNips2016} & 613.00 &28.50& 9.08& 0.71 &0.80\\
\hline 
{\textbf{Our work}} & 1344.40 & \textbf{39.80} & 13.12 & 0.70 & 0.79\\
\hline 
\end{tabular}}
\label{table_open_ended_evaluation}
\vspace{-6mm}
\end{center}
\end{table}
Although on average, BoW and Local-LDA achieved better performance than our approach, the difference is minor, and the discriminative power of those approaches is lower. This is expected since the BoW and Local-LDA learned fewer categories, and it is easier to get better classification performance in fewer categories. 

\subsection {System Demonstration}

\begin{wrapfigure}{r}{0.5\linewidth}
\vspace{-10mm}
\begin{center}
\resizebox{\linewidth}{!}{
  \includegraphics[width=0.45\linewidth, trim= 0cm 1cm 0.1cm 0cm,clip=true]{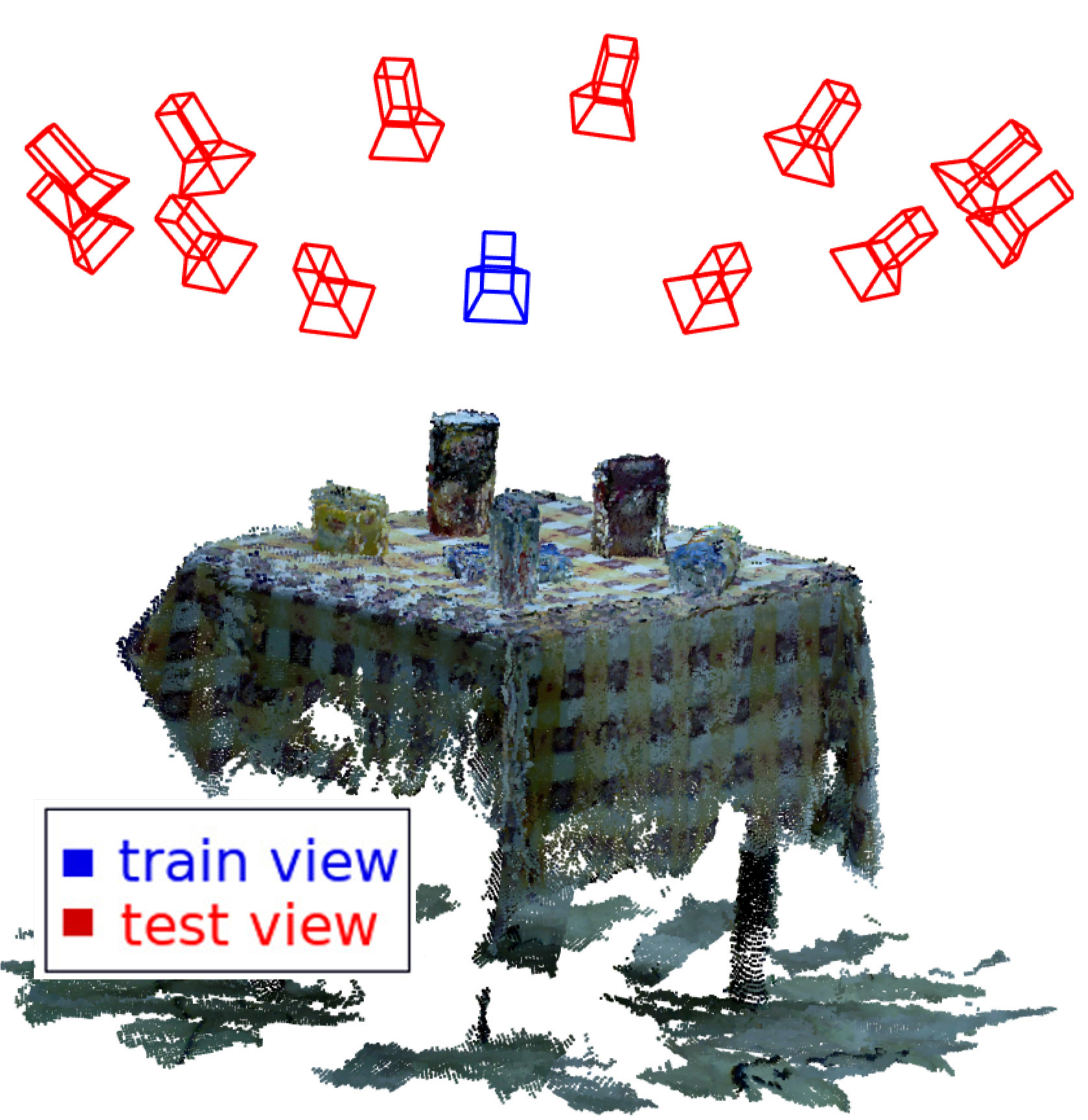} 
}
\caption{ Experimental setup using Imperial College dataset~\cite{doumanoglou2016recovering}.}
\vspace{-5mm}
\label{fig:real_demo}
\end{center}
\end{wrapfigure}
In this demonstration, the system initially had no prior knowledge, and all objects are recognized as \emph{Unknown}. Later, a user interacts with the system and teaches all object categories including \emph{amita}, \emph{colgate}, \emph{lipton}, \emph{elite}, \emph{oreo} and \emph{softkings} to the system using the objects extracted from a scene captured from the blue camera pose as shown in Fig.~\ref{fig:real_demo} (\emph{left}). The system conceptualizes those categories using the extracted object views. Afterwards, the system is tested by the remaining 12 scenes captured from different viewpoints (i.e., shown by red cameras). The system could recognize all objects properly by using the knowledge learned from the first scene. Some misclassifications also occurred throughout the demonstration. The underlying reason was that, at some points, the object tracking could not track the object properly and the distinctive parts of the object were not included in the object's point cloud. This evaluation illustrates the process of learning object categories in an open-ended fashion. A video of this demonstration is available online at: \href{https://youtu.be/zjucGaAwnTE}{\small \cblue{https://youtu.be/zjucGaAwnTE}}


\section{Conclusion}
\label{sec:conclusion}
In this paper, we have tackled the problem of open-ended object category learning and recognition by proposing a new object representation, which is suitable for both fine-grained and basic-level object categorization. In particular, we described each object based on a set of \emph{general latent visual topics} and a set of \emph{category-specific} dictionaries. An extensive set of experiments was carried out to assess the performance of the proposed approach. Experimental results show that the overall classification performance obtained with the proposed approach is clearly better than the best accomplishments achieved with the state-of-the-art methods. Moreover, the best scalability was obtained with the proposed approach, followed by the Local-LDA~\cite{kasaeiNips2016}. Concerning computational time, the best result was obtained with BoW~\cite{kasaei2018towards}, immediately followed by the Open-Ended LDA approach~\cite{blei2003latent}. A real demonstration proved that the agent could learn new categories from very few examples in an incremental and open-ended manner.


{\small
\bibliographystyle{IEEEtran}
\bibliography{bibs/refs.bib,bibs/refs_IROS18.bib}
}

\end{document}